\documentclass[10pt,twocolumn,letterpaper]{article}
\pdfoutput=1

\usepackage{iccv}
\usepackage{times}
\usepackage{epsfig}
\usepackage{graphicx}
\usepackage{amsmath}
\usepackage{amssymb}
\usepackage{bbm}
\usepackage{subcaption}
\usepackage{multirow}
\usepackage{adjustbox}
\usepackage[dvipsnames]{xcolor}
\usepackage{tabu}
\usepackage{floatrow}
\usepackage{booktabs}
\usepackage{makecell}
\usepackage{epigraph}
\usepackage{mathtools}

\usepackage{algorithm}
\usepackage{algorithmic}

\usepackage{animate}

\usepackage[percent]{overpic}

\usepackage[pagebackref=true,breaklinks=true,letterpaper=true,colorlinks,bookmarks=false]{hyperref}

\newtheorem{assumption}{Assumption}

\newcommand{\ra}[1]{\renewcommand{\arraystretch}{#1}}

\newcommand*{\ShowNotes}{}

\definecolor{darkred}{rgb}{0.7,0.1,0.1}
\definecolor{darkgreen}{rgb}{0.1,0.7,0.1}
\definecolor{cyan}{rgb}{0.7,0.0,0.7}
\definecolor{dblue}{rgb}{0.2,0.2,0.8}
\definecolor{maroon}{rgb}{0.76,.13,.28}
\definecolor{burntorange}{rgb}{0.81,.33,0}

\ifdefined\ShowNotes
  \newcommand{\colornote}[3]{{\color{#1}\bf{#2: #3}\normalfont}}
\else
  \newcommand{\colornote}[3]{}
\fi

\iccvfinalcopy % *** Uncomment this line for the final submission

\def\iccvPaperID{892} % *** Enter the ICCV Paper ID here

\ificcvfinal\fi
\begin{document}

%%%%%%%%% TITLE
\title{Dynamic-Net: Tuning the Objective Without Re-training for Synthesis Tasks}
% \author{Alon Shoshan\qquad Roey Mechrez \qquad Lihi Zelnik-Manor\\
% Technion - Israel Institute of Technology\\
% {\tt\small \{shoshan@campus,roey@campus,lihi@ee\}.technion.ac.il}}
\author{Alon Shoshan\\
Technion, Israel\\
{\tt\small shoshan@campus.technion.ac.il}
\and
Roey Mechrez\\
Technion, Israel\\
{\tt\small roey@campus.technion.ac.il}
\and
Lihi Zelnik-Manor\\
Technion \& Alibaba Group\\
{\tt\small lihi@technion.ac.il}
}

% \author[1]{Alon Shoshan}
% \author[1]{Roey Mechrez}
% \author[1,2]{Lihi Zelnik-Manor}
% \affil[1]{Technion, Israel}
% \affil[2]{Alibaba Group}

%\maketitle
%\ificcvfinal\thispagestyle{empty}\fi

%%%%%%%%%%%%%%%%  teaser figure
% GIF: for arXiv
\twocolumn[{%
\renewcommand\twocolumn[1][]{#1}%
\maketitle
\ificcvfinal\thispagestyle{empty}\fi
\begin{center}
\centering
%\animategraphics[autoplay,loop,width=\linewidth]{1}{images/teaser_gif/teaser_v7_}{0}{5}
\animategraphics[autoplay,loop,width=\linewidth]{1}{images/teaser_gif/teaser_v8_}{0}{5}
\captionof{figure}{\textbf{Dynamic-Net:} We propose an approach that enables traversing the ``objective-space'', spanned by two different objectives, at test-time, without re-training, as illustrated by the blue dot moving along the blue curve in the plot.
This is different from the common practice of training a separate network for each objective, represented by $\times$'s on the plot.
Using a single Dynamic-Net we can tune the level of stylization of an image, monitor completion quality per image, or control facial attributes, all interactively at test-time, without re-training. [Animated figure, please view in Acrobat Reader].
}
\label{fig:teaser}
\end{center}
}]
% STATIC for ICCV camera-ready
%\input{figures/fig_static_teaser.tex}

%%%%%%%%% ABSTRACT
\begin{abstract}
One of the key ingredients for successful optimization of modern CNNs is identifying a suitable objective. To date, the objective is fixed a-priori at training time, and any variation to it requires re-training a new network. In this paper we present a first attempt at alleviating the need for re-training. Rather than fixing the network at training time, we train a ``Dynamic-Net'' that can be modified at inference time. Our approach considers an ``objective-space'' as the space of all linear combinations of two objectives, and the Dynamic-Net is emulating the traversing of this objective-space at test-time, without any further training. We show that this upgrades pre-trained networks by providing an out-of-learning extension, while maintaining the performance quality. The solution we propose is fast and allows a user to interactively modify the network, in real-time, in order to obtain the result he/she desires. We show the benefits of such an approach via several different applications. %Our code is available at \url{https://github.com/AlonShoshan10/dynamic_net}.
\vfill\null
\end{abstract}

%%%%%%%%% BODY TEXT
\vspace{-2.1cm}
\section{Introduction}
\label{sec:intro}
\vspace{-0.1cm}
\noindent\emph{``I can't change the direction of the wind, but I can adjust my sails to always reach my destination''}

\hfill \textemdash Jimmy Dean
\\

\vspace{-0.1cm}
%%%%%%%%%%%%%%%%%%%%%%%%%\paragraph{Current sota}
A common practice in image generation is to train a deep network with an appropriate objective.
The objective is often complex and integrates multiple loss terms, e.g., in style transfer~\cite{gatys2016image,johnson2016perceptual}, super-resolution~\cite{blau2018perception,mechrez2018maintaining}, inpainting~\cite{liu2018image}, image-to-image transformations~\cite{isola2017image}, domain transfer~\cite{mechrez2018contextual,kim2017learning} and attribute manipulation~\cite{shen2017learning}. 
%
%
% For example, in style transfer~\cite{gatys2016image,johnson2016perceptual} the objective integrates between similarity to the input image and similarity to the style image.
% In super-resolution, as was shown in \cite{blau2018perception}, the objective trade-offs low PSNR and high perceptual quality. 
% In fact, a trade-off between multiple loss terms is manifested in many computer vision applications, such as compression, denoising and inpainting~\cite{liu2018image}, image-to-image transformations~\cite{isola2017image}, domain transfer~\cite{mechrez2018contextual,kim2017learning} and attribute manipulation~\cite{shen2017learning}. 
% It is even relevant to 3D related tasks such as normal and depth estimation~\cite{mechrez2018learning} where the goal is to achieve realistic rendering. 
%
%
%%%%%%%%%%%%%%%%%%%%%%%%%\paragraph{The challenge}
To date, the choice of the specific objective, and the trade-off between its multiple terms, are set a-priori during training. 
This results in trained networks that are fixed for a specific working point.
%, as illustrated in Figure~\ref{fig:teaser}.
This is limiting for three reasons.
First, oftentimes one would like the flexibility to produce different results, e.g., stronger or weaker style transfer. 
Second, in many cases the best working point is different for different inputs.
Last, it is hard to predict the optimal working point, especially when the full objective is complex and when adversarial training~\cite{goodfellow2014generative} is incorporated.
Therefore, practitioners perform greedy search over the space of objectives during training, which demands significant compute time.
%Attempts for automatic hyper-parameter optimization have also been made \cite{li2017hyperband}, but these are still preliminary and no generic solutions are available. Furthermore, we are not familiar with works that optimize the objective itself.

% %%%%%%%%%%%%%%%%%%%%%%%%%\paragraph{Research question}
% This raises an important question: \textit{How can we control the compromise between the two objectives?} The importance of this question is two-fold: first, the optimal working point along the objective curve (illustrate in Figure~\ref{fig:dynamic_alg_example}) is domain specific and moreover it is image specific. 

%%%%%%%%%%%%%%%%%%%%%%%%%\paragraph{Key idea - Proposed approach}
In this paper we propose an alternative approach, called \emph{Dynamic-Net}, that resolves this for some scenarios.
Rather than training a single fixed network, we split the training into two phases.
In the first, we train the blocks of a ``main network'' using a certain objective.
At the second phase we train additional residual~\cite{He2016DeepRL} ``tuning-blocks'', using a different objective.
Then, at inference time, we can decide whether we want to incorporate the tuning-blocks or not and even control their contribution.
This way, we actually have at hand a dynamic network that can be assembled at inference time from the main network and tuning-blocks.
Our underlying assumption is that the tuning-blocks can capture the variation between the two objectives, thus allowing traversal of the objective space.
The Dynamic-Net can thus be easily geared towards the first or second objective, by tuning scalar parameters, at test-time.

%%%%%%%%%%%%%%%%%%%%% {key idea}
The key idea behind our approach is inspired by the Jimmy Dean citation at the beginning of the introduction.
We acknowledge that we cannot directly modify the objective at test-time.
However, what we can do is modify the latent space representation. 
Therefore, our approach relies on manipulation of deep features in order to emulate a manipulation in objective space.

%%%%%%%%%%%%%%%%%%%%%%%%%\paragraph{Contribution}
The main advantages of the Dynamic-Net are three-fold.
First, using a single training session the Dynamic-Net can emulate networks trained with a variety of different objectives, for example, networks which produce stronger or weaker stylization effects, as illustrated in Figure~\ref{fig:teaser}.
Second, it facilitates image-specific and user-specific adaptation, without re-training. 
Via a simple interface, a user can interactively choose at real-time the level of stylization or a preferred inpainting result.
Last, the ability to traverse the objective space at test-time shrinks the search-space during training.
More specifically, we show that even when the choice of objective for training is sub-optimal, the Dynamic-Net can reach a better working point. 

% This is useful when the tuning-blocks capture something that is meaningful to the user. 
% For example, the tuning blocks can be trained to change the style of an image and at inference time the user can adaptively control the level of stylization.

% The second benefit is removing the need for hyper-parameter tuning during training. \roey{too general}
% Rather than seeking the best objective during training, which is time consuming, the Dynamic-Net enables real-time tuning of the network to yield a desired working points.

%%%%%%%%%%%%%%%%%%%%%%%%%\paragraph{Empirical contribution}
We show these benefits through a broad range of applications in image generation, manipulation and reconstruction. 
We explore a variety of objectives and architectures, and present both qualitative and quantitative evaluation. Our code is available at \url{https://github.com/AlonShoshan10/dynamic_net}.

\section{Related Work}
\label{sec:related-work}
\vspace{-0.1cm}
\paragraph{Multi-loss objectives}
Many state-of-the-art solutions for image manipulation, generation and reconstruction utilize a multi-loss objective. 
For example, Isola et al.~\cite{isola2017image} combine $L1$ and adversarial loss~\cite{goodfellow2014generative} for image-to-image transformation. Johnson et al.~\cite{johnson2016perceptual} trade-off between a style loss (i.e. Gram loss) and content loss (i.e. the perceptual loss) for fast style-transfer and super-resolution, SRGAN~\cite{ledig2017photo} balances between a content loss and adversarial loss for perceptual super-resolution, and \cite{zhou2018non} combines $L1$, a style loss and an adversarial loss for texture synthesis.
In all of these cases, the weighting between the loss terms is fixed during training, producing a trained network that operates at a specific working point.

The impact of the trade-off between multiple objectives has been discussed before in several contexts. \cite{blau2018perception} show that in image restoration algorithms there is an inherent trade-off between distortion and perceptual quality. They analyze the trade-off and show the benefits of different working points.
In~\cite{gatys2016image} it is shown empirically that a different balance between the style loss and content loss leads to different stylization effects.

The importance and difficulty of choosing the optimal balance between different loss terms is tackled by methods for multi-task learning~\cite{chen2017gradnorm,misra2016cross,li2017learning,ruder2017overview,rosenbaum2017routing}.
In these works a variety of solutions have been proposed for learning the weights for different tasks.
Mutual to all of these methods is that their outcome is a network trained with a certain fixed balance between the objectives.
\vspace{-0.1cm}

\paragraph{Deep feature manipulation}
The approach we propose is based on training ``tuning-blocks'' that learn how to manipulate deep features in order to achieve a certain balance between the multiple objectives.
It is thus related to methods that employ manipulation of deep features in latent space.
These methods are based on the basic hypothesis of \cite{bengio2013better} that ``CNNs linearize the manifold of natural images into a Euclidean subspace of deep features'', suggesting that linear interpolation of deep features makes sense.
Inspired by this hypothesis, \cite{upchurch2017deep} learn the linear ``direction'' that modifies facial attributes, such as adding glasses or a mustache.
In \cite{chen2018facelet} a more sophisticated manipulation approach is proposed. 
They introduced blocks to be added to an auto-encoder network in order to learn the required manipulation to modify a facial attribute. 
While producing great results on manipulation of face images, their approach implicitly assumes that the training images are similar and roughly aligned. %\alon{need to go over this.}
%the input and manipulated images are similar and roughly aligned.

% \begin{figure}
% \includegraphics[width=\linewidth]{images/single_block.png}
% \caption{\textbf{Single-block framework:} Our training has two steps: 
% (i) First the ``main'' network $\theta$ (green blocks) is trained to minimize ${\cal O}_0$. 
% (ii) Then $\theta$ is fixed, a tuning block $\psi$ is added (orange block), and trained to minimize ${\cal O}_1$. 
% The output $\hat y_1$ approximates the output $y_1$ one would get from training the main network $\theta$ with objective ${\cal O}_1$. 
% At test-time, we can emulate results equivalent to a network trained with objective ${\cal O}_m$ by tuning the parameter $\alpha_m$ that determines the latent representation $z_m$.
% }
% \label{fig:single_block}
% \end{figure}
\begin{figure*}
\centering
\begin{tabular}{cc}
\includegraphics[height=4.1cm,keepaspectratio]{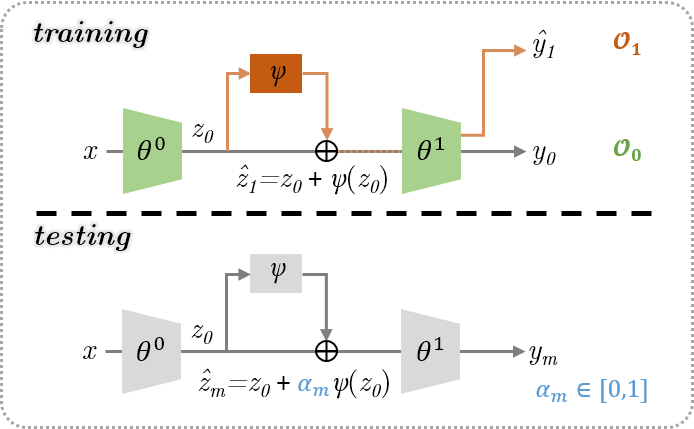}&
\includegraphics[height=4.1cm,keepaspectratio]{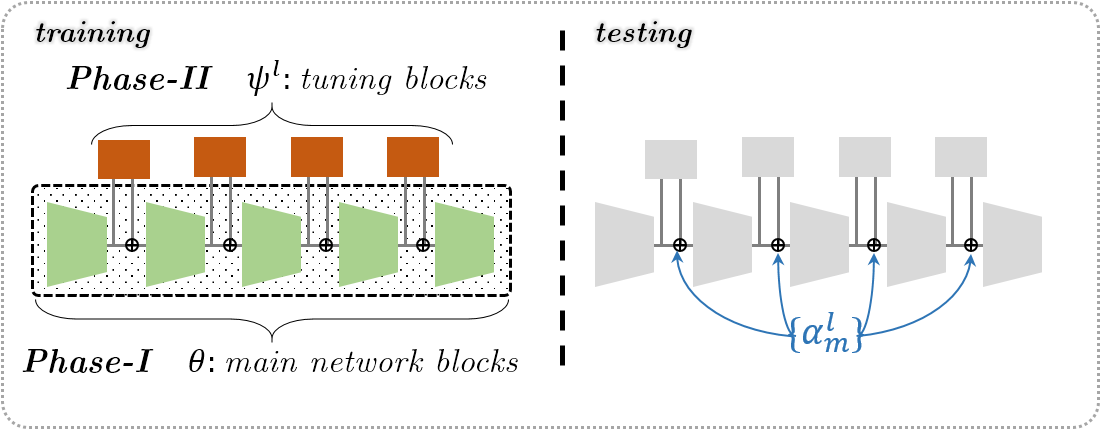}\\
(a) Single-block framework & (b) Multi-block framework
\end{tabular}
\vspace{-0.1cm}
\caption{\textbf{Proposed framework:} Our training has two steps: 
(i) First the ``main'' network $\theta$ (green blocks) is trained to minimize ${\cal O}_0$. 
(ii) Then $\theta$ is fixed, one or more tuning-blocks $\psi$ are added (orange block), and trained to minimize ${\cal O}_1$. 
The output $\hat y_1$ approximates the output $y_1$ one would get from training the main network $\theta$ with objective ${\cal O}_1$. 
At test-time, we can emulate results equivalent to a network trained with objective ${\cal O}_m$ by tuning the parameter $\alpha_m$ (in blue) that determines the latent representation $z_m$. Our method can be applied as (a) a single-block framework or as (b) multi-block framework. 
}
\label{fig:single_block}
\end{figure*}
\vspace{-0.1cm}
\section{Proposed Approach:  Dynamic-Net}
\label{sec:proposed-approach}
\vspace{-0.1cm}

%We start by describing a solution that uses a single tuning parameter, and then extend to a more generic one with multiple parameters.

To date, one has to re-train the network for each objective.
In this section we propose Dynamic-Net that allows changing the objective at inference time, without re-training.
Dynamic-Nets can emulate a plethora of ``intermediate working points'' between two given objectives ${\cal O}_0$ and ${\cal O}_1$, by simply tuning a single parameter.
One can think of this as implicit interpolation between the objectives.
Our solution is relevant in cases where such an interpolation between the objectives ${\cal O}_0$ and ${\cal O}_1$ is meaningful.

% Schematically, our pipeline first trains the \emph{main-blocks} of the network with objective ${\cal O}_0$ and then trains additional \emph{tuning-blocks} with objective ${\cal O}_1$.
% At inference time we integrate between the main-blocks and tuning-blocks.
% By modifying a parameter the integrated network can approximate ``intermediate working points'' ${\cal O}_m$  between ${\cal O}_0$ and ${\cal O}_1$.
% The notion of ``intermediate working point'' is not well defined, and hence it is discussed below.

%\paragraph{Intermediate working points:}
To provide some intuition we begin with an example.
A common scenario where an ``intermediate working point'' is intuitive, is when the objectives consist of a super-position of two loss terms:
${\cal O}_0\!=\!{\cal L_A}\!+\!\lambda_{0} {\cal L_B}$ and 
${\cal O}_1\!=\!{\cal L_A}\!+\!\lambda_{1} {\cal L_B}$,
where $\cal{L_A}$, $\cal{L_B}$ are loss terms, and $\lambda_0,\lambda_1$ are scalars. 
Assuming, without loss of generality, that $\lambda_{0}\!\leq\!\lambda_{1}$, an intermediate working point corresponds to an objective ${\cal O}_m\!=\!{\cal L_A}\!+\!\lambda_{m} {\cal L_B}$, such that $\lambda_{0}\!\leq\!\lambda_{m}\!\leq\!\lambda_{1}$.
Our goal is to approximate at inference time the results of a network trained with any objective ${\cal O}_m$, while using only ${\cal O}_0$ and ${\cal O}_1$ during training.

% A network can be trained for different working points by tuning the parameter $\lambda$.
% Training with objective ${\cal O}_0\!=\!{\cal L_A}\!+\!\lambda_{0} {\cal L_B}$ leads to output $y_0$ while training with objective ${\cal O}_1\!=\!{\cal L_A}\!+\!\lambda_{1} {\cal L_B}$ leads to output $y_1$.

% To clarify what we mean by ``intermediate working points'' we assume, without loss of generality, that 
% $\lambda_{0}\!\leq\!\lambda_{1}$.
% An intermediate working point corresponds to objective ${\cal O}_m\!=\!{\cal L_A}\!+\!\lambda_{m} {\cal L_B}$, such that $\lambda_{0}\!\leq\!\lambda_{m}\!\leq\!\lambda_{1}$.
% output $y_m$ and latent representation $z_m$.

%For simplicity of presentation we start with a simplified setup where we add a single tuning block, at a single layer of the network. Later on we extend to multiple tuning blocks at multiple different layers of the network.

The key idea behind the approach we propose, is to use interpolation in latent space in order to approximate the intermediate objectives. 
For simplicity of presentation we start with a simple setup that uses linear interpolation, at a single layer of the network. 
Later on we extend to non-linear interpolation.

%%%%%%%%%%%%%%%%%%%%%%%%%%%%%%%%%%%%%%%%%%%%%%%%%%%%%%%%%%%%%%%%%%%%%%%%%%%%
\subsection{Single-block Dynamic-Net}
\vspace{-0.1cm}
Our single-block framework is illustrated in Figure~\ref{fig:single_block}(a). 
It first trains a CNN, to which we refer as the ``\emph{main network blocks}'', with objective ${\cal O}_0$.
We then add an additional block to the network, to which we refer as the ``\emph{tuning-block}'' $\psi$ that learns the ``direction of change'' in latent space $z$, that corresponds to shifting the objective from ${\cal O}_0$ to another working point ${\cal O}_1$.
Our hypothesis is that walking along the ``direction of change'' in latent space can emulate a plethora of ``intermediate'' working points ${\cal O}_m$ between ${\cal O}_0$ and ${\cal O}_1$.

In further detail, our pipeline is as follows:

%%      \paragraph{Algorithm outline:}
\noindent $\triangleright$  \underline{Training:}
\begin{itemize}
\item Train the main network blocks by setting the objective to ${\cal O}_0$.
\item Fix the values of the main network, add a tuning-block $\psi$ between layers $l$ and $l\!+\!1$, and post-train only $\psi$ by setting the objective to ${\cal O}_1$.
The block $\psi$ will capture the variation between the latent representations $z_0$ and $z_1$, that correspond to ${\cal O}_0$ and ${\cal O}_1$, respectively. 
\end{itemize}

\noindent $\triangleright$ \underline{Testing:}\\
Fix both the main blocks as well as the tuning block $\psi$, and do as follows:
\begin{itemize}
\item Propagate the input until layer $l$ of the main network to get $z_0$. 
\item Generate an ``intermediate'' point in latent space, $z_m=z_0+\alpha_m \psi(z_0)$, by tuning the scalar parameter $\alpha_m$. 
\item Propagate $z_m$ through the rest of the main network to obtain outcome $y_m$ that corresponds to objective ${\cal O}_m$.
\end{itemize}

The justification for our approach stems from the following two assumptions:

\begin{assumption}\label{as:1}
We adopt the hypothesis of~\cite{bengio2013better} that ``CNNs linearize the manifold of natural images into a Euclidean subspace of deep features''.
\end{assumption}
\noindent This assumption implies that the latent representation of an intermediate point can be written as
$z_m\!=\!z_0 + \alpha_m(z_1-z_0)$ where $\alpha_m\!\in\![0,1]$.
Setting $\alpha_m\!=\!0$ yields working point 0 while setting $\alpha_m\!=\!1$ yields working point 1.

\begin{assumption}\label{as:2}
For any pair of working points  ${\cal O}_0$,${\cal O}_1$, with corresponding latent representations $z_0,z_1$, it is possible to train a block $\psi$ such that $z_1 \approx z_0+\psi(z_0)$.
\end{assumption}
\noindent Putting assumptions~\ref{as:1} and~\ref{as:2} together suggests that we can approximate any intermediate working point ${\cal O}_m$ by computing
$\hat{z}_m = z_0+\alpha_m \psi(z_0)$ and have that $z_m \approx \hat{z}_m$.
%

%\paragraph{Interpolation in latent vs. objective space:}
To provide further intuition we revisit the example where the objectives are of the form  
${\cal O}\!=\!{\cal L_A}\!+\!\lambda {\cal L_B}$.
Here the parameter $\lambda$ controls the balance between the two loss terms ${\cal L_A}$ and ${\cal L_B}$. 
To interpolate in objective space we would like to modify $\lambda$ but this is not possible to do directly at test-time.
Instead, our scheme enables interpolation in latent space by modifying the parameter $\alpha$, which controls $z_m$.
Our main hypothesis is that the suggested training scheme will lead to a proportional relation between $\alpha$ and $\lambda$. 
That is, increasing $\alpha$ will correspond to a monotonic increase in $\lambda$, thus implicitly achieving the desired interpolation in objective space.
%This idea is presented schematically in Figure~\ref{fig:single_block} where we show the training and inference pipelines as well as the relation between the objectives, the outputs, and the latent space. 

In the more general case, when the objectives ${\cal O}_0$ and ${\cal O}_1$ are of different forms, the interpolation we propose in objective space cannot be formulated mathematically so intuitively. 
Nonetheless, the conceptual meaning of such interpolation could be sensible. 
For example, we could train an image generation network with two different adversarial objectives, one that prefers blond hair and another that prefers dark hair. 
Interpolating between the two objectives should correspond to generating images with varying hair shades. 
Therefore, to prove broad applicability of the proposed approach to a variety of objectives, we present in Section~\ref{sec:experiments} several applications and corresponding results.

%%%%%%%%%%%%%%%%%%%%%%%%%%%%%%%%%%%%%%%%%%%%%%%%%%%%%%%%%%%%%%%%%%%%%%%%%%%%%%%%%%%%%%%%%%%%%%%%%%%%%
\subsection{Multi-block Dynamic-Net}
\vspace{-0.1cm}
In practice, adding a single tuning-block, at a specific layer, might be insufficient.
It limits the manipulation to linear transformations in a single layer of the latent space.
Therefore, we propose adding multiple blocks, at different layers of the network as illustrated in Figure~\ref{fig:single_block}(b).
% (i) ``Cascade Blocks'' are added in a cascade, in between layers of the network. (ii) ``Parallel Blocks'' are appropriate for Encoder-Decoder architectures as the added blocks are placed on skip-connections between corresponding layers of the encoder and decoder.

The training framework is similar to that of single-block, except that now we have multiple tuning blocks $\psi^l$, each associated with a corresponding weight $\alpha_m^l$. When training the tuning-blocks we fix all the weights to $\alpha_m^l=1$. Then at inference-time, we can tune each of the weights independently to yield a plethora of networks and results.

\section{Experiments}
\label{sec:experiments}
\vspace{-0.1cm}

In this section we present experiments with several applications that demonstrate the utility of the proposed Dynamic-Net and support the validity of our hypotheses. 
In order to emphasize broad applicability we selected applications of varying nature, with a variety of loss functions and network architectures, as summarized in Table~\ref{tab:applications}.
Tuning-blocks were implemented as $ conv\!-\!relu\!-\!conv\!-\!relu\!-\!conv $. Further implementation details, architectures and parameter values are listed in the supplementary.

The motivation behind Dynamic-Net was three-fold: (i) provide ability to modify the working point at test-time, (ii) allow image-specific adaptation, and (iii) reduce the dependence on optimal objective selection at training time. 
In what follows we explore these contributions one by one, through various applications.

In the next subsections, if not stated otherwise, we used the multi-block framework while setting all $\{\alpha^l\}$ to be equal, i.e., $\alpha^0=\alpha^1...\equiv\alpha$, and we tune $\alpha$.

%%%%%%%%%%%%%%%%%%%%%%%%%%%%%%%%%%%%%%%%%%%%%%%%%%%%%%%%%%%%%%%%%%%%%%%%%%%%%%
%\subsection{Method Generality:}
%\todo{\textbf{Generality:} 5 applications, 4 architectures, 4 types of loss terms. Two types of objective space: (i) trade-off / super-positions, (ii) two different objectives.}
% Recall that our main hypothesis is that changing the value of $\alpha$ and modify the latent representation will lead to a proportional change in the working point, as if, we could re-train the net with a different objective (i.e. change $\lambda$). In following section we empirically confirm the hypothesis. This is done by an extensive experiment on five different application and a variety of loss terms. 

\begin{table}
    \setlength{\tabcolsep}{2pt}
    \centering
	\ra{1}
	\begin{tabu}{@{}lcccc@{}}
    \toprule
	\textbf{Application} & \phantom{ab} &  \textbf{Objectives} & \phantom{ab} & \textbf{Architecture}\\
    \midrule
        Style Transfer && ${\cal L}_{content},{\cal L}_{style}$ &&  \cite{johnson2016perceptual}\\
        %Image-to-Image && ${\cal L}_{L1},{\cal L}_{adv}$ &&  \cite{isola2017image}\\
        Image Completion && ${\cal L}_{L1},{\cal L}_{adv}$ &&  \cite{isola2017image}\\
        %Super-Resolution && ${\cal L}_{adv},{\cal L}_{per},{\cal L}_{L1}$ &&  \cite{wang2018esrgan}\\
        Face Generation && ${\cal L}_{adv}$ &&  \cite{upchurch2017deep}\\
    \bottomrule
    \end{tabu}
    \caption{\textbf{Applications summary:} We evaluate Dynamic-Net on three different applications: image manipulation (style transfer), reconstruction (image completion) and generation (faces). The applications minimize different loss terms and are based on a variety of architectures.  
    }
    \label{tab:applications}
\end{table}

% In Style Transfer we allow a user to interactively add/remove/modify the style smoothly. In Edges2Image we tune the network per image, leading to overall more perceptually realistic results.Finally, we show that we can synthesize faces using a single network, yet controlling their attributes, e.g, we can choose any hair shade from dark to blond. 

%%%%%%%%%%%%%%%%%%%%%%%%%%%%%%%%%%%%%%%%%%%%%%%%%%%%%%%%%%%%5555
\subsection{Tuning the objective at test-time\\:: Style Transfer}
\label{sec:style_transfer}
\vspace{-0.1cm}
%\todo{\textbf{Tuning the objective at inference time:} all applications, style transfer, extrapolation }

Our first step is to show that the proposed approach can indeed traverse the objective space, and emulate multiple meaningful working points at test-time, without re-training. 
We chose to show this via experiments in Style Transfer.

\begin{figure*}[t]
\centering
\centering
\setlength{\tabcolsep}{1pt}

\definecolor{fixed_nets}{RGB}{192,80,77}
\definecolor{ours}{RGB}{79,129,189}
\definecolor{base_line}{RGB}{155,187,89}

\fboxsep=0mm%padding thickness
\fboxrule=1pt%border thickness

\setlength{\tabcolsep}{0pt}
\begin{tabular}{c c c c c}
    $\lambda\!=\!2\cdot10^4$ & 
    $\lambda\!=\!5\cdot10^4$ &
    $\lambda\!=\!10^5$ &
    %$\lambda\!=\!8\cdot10^4$ &
    $\lambda\!=\!2\cdot10^5$ &
    $\lambda\!=\!5\cdot10^5$
    \\
    \rotatebox{90}{\hspace{9.5mm}\tiny{Fixed nets \textcolor{white}{p}}}
    \fcolorbox{fixed_nets}{fixed_nets}{\includegraphics[width=0.195\linewidth]{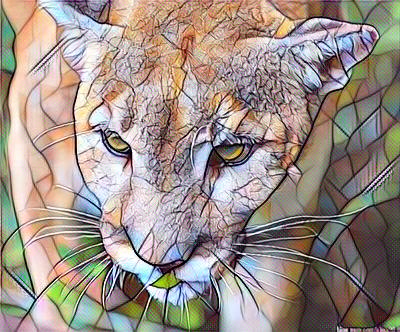}} &
    %\fcolorbox{fixed_nets}{fixed_nets}{\includegraphics[width=0.195\linewidth]{style_transfer_images/graph_images/000001_3e4_400.jpg}} &
    \fcolorbox{fixed_nets}{fixed_nets}{\includegraphics[width=0.195\linewidth]{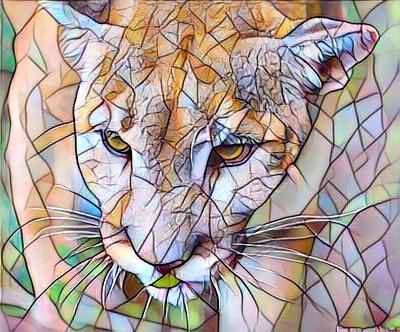}} &
    \fcolorbox{fixed_nets}{fixed_nets}{\includegraphics[width=0.195\linewidth]{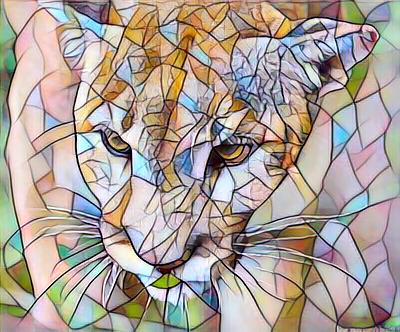}} &
    %\fcolorbox{fixed_nets}{fixed_nets}{\includegraphics[width=0.141\linewidth]{style_transfer_images/graph_images/000003_8e4_400.jpg}} &
    \fcolorbox{fixed_nets}{fixed_nets}{\includegraphics[width=0.195\linewidth]{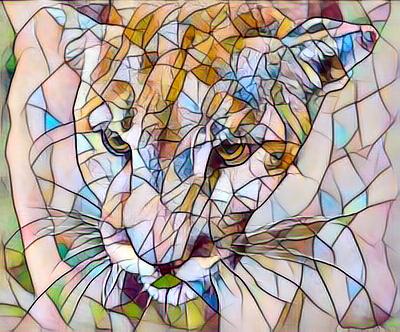}} &
    \fcolorbox{fixed_nets}{fixed_nets}{\includegraphics[width=0.195\linewidth]{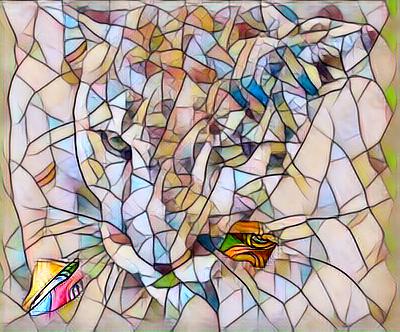}}
    \\
    $\alpha\!=\!0$ & 
    %$\alpha\!=\!0.2$ &
    $\alpha\!=\!0.25$ &
    %$\alpha\!=\!0.35$ &
    $\alpha\!=\!0.5$ &
    $\alpha\!=\!0.75$ &
    $\alpha\!=\!1$
    \\
    \rotatebox{90}{\hspace{9mm}\tiny{Image interp}}
    \fcolorbox{base_line}{base_line}{\includegraphics[width=0.195\linewidth]{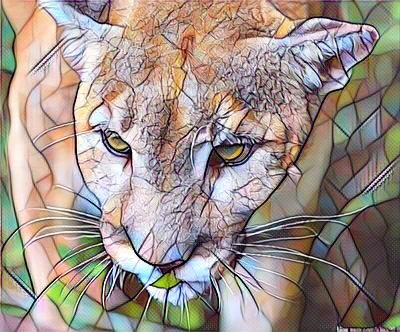}} &
    \fcolorbox{base_line}{base_line}{\includegraphics[width=0.195\linewidth]{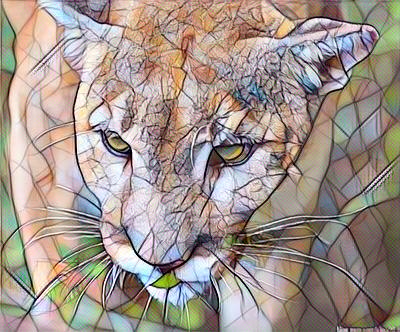}} &
    %\fcolorbox{base_line}{base_line}{\includegraphics[width=0.141\linewidth]{style_transfer_images/graph_images/000000_2e4_to_5e5_lioness_400_0350.jpg}} &
    \fcolorbox{base_line}{base_line}{\includegraphics[width=0.195\linewidth]{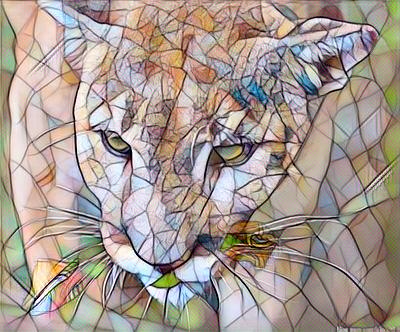}} &
    \fcolorbox{base_line}{base_line}{\includegraphics[width=0.195\linewidth]{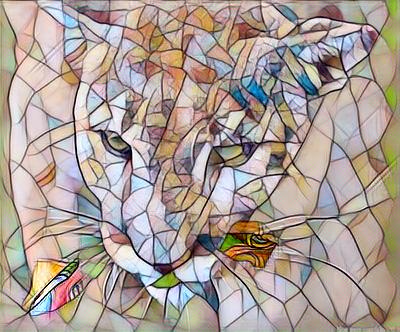}} &
    \fcolorbox{base_line}{base_line}{\includegraphics[width=0.195\linewidth]{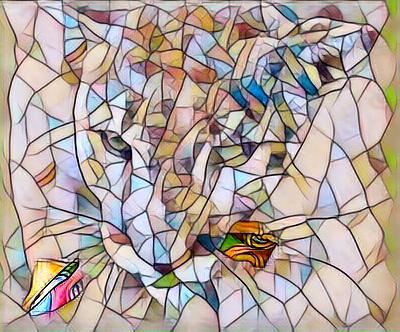}}
    \\
    \rotatebox{90}{\hspace{11mm}\tiny{Ours \textcolor{white}{p}}}
    \fcolorbox{ours}{ours}{\includegraphics[width=0.195\linewidth]{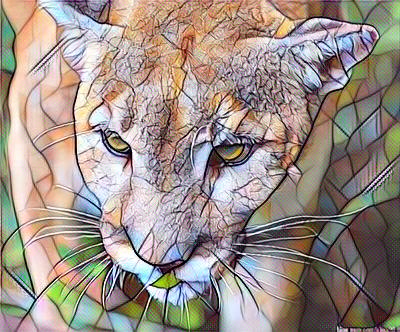}} &
    \fcolorbox{ours}{ours}{\includegraphics[width=0.195\linewidth]{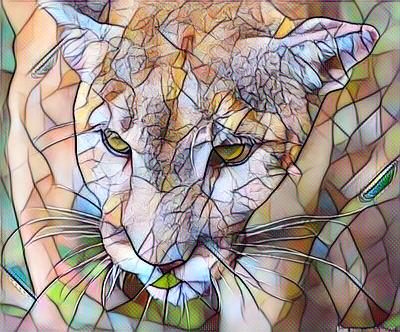}} &
    %\fcolorbox{ours}{ours}{\includegraphics[width=0.141\linewidth]{style_transfer_images/graph_images/A_400_0350.jpg}} &
    \fcolorbox{ours}{ours}{\includegraphics[width=0.195\linewidth]{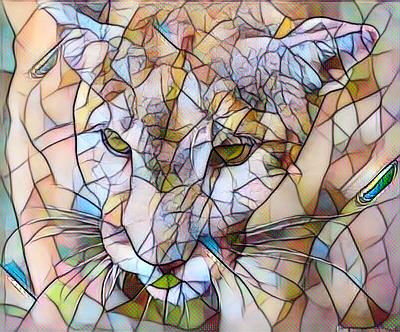}} &
    \fcolorbox{ours}{ours}{\includegraphics[width=0.195\linewidth]{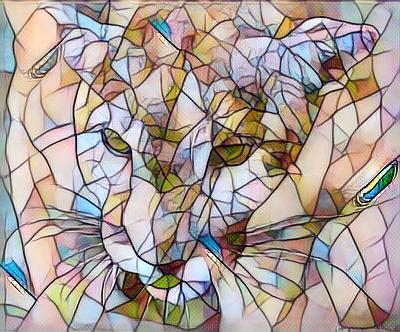}} &
    \fcolorbox{ours}{ours}{\includegraphics[width=0.195\linewidth]{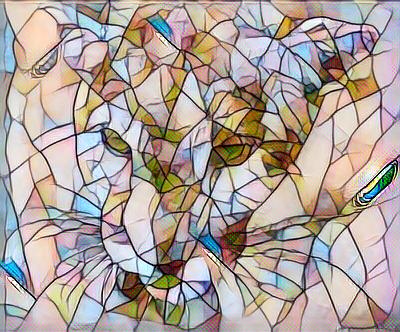}}
    \\
\end{tabular}

\setlength{\tabcolsep}{4pt}
\begin{tabular}{c c c}
    \multirow{2}{*}{\includegraphics[width=0.4\linewidth]{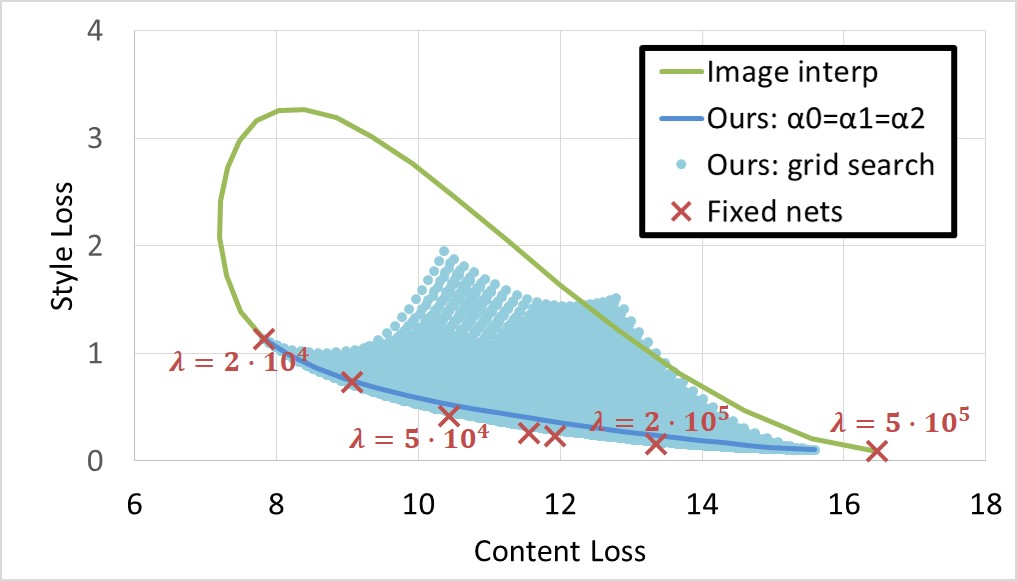}} &
    Image interp & Ours
    \\
    &
    %\fcolorbox{base_line}{base_line}{\includegraphics[width=0.22\linewidth]{style_transfer_images/graph_images/4.png}} &
    %\fcolorbox{ours}{ours}{\includegraphics[width=0.22\linewidth]{style_transfer_images/graph_images/5.png}}
    \fcolorbox{base_line}{base_line}{\includegraphics[trim={20px 32px 200px 120px}, clip, width=0.2\linewidth]{style_transfer_images/graph_images/000000_2e4_to_5e5_lioness_400_0250.jpg}} &
    \fcolorbox{ours}{ours}{\includegraphics[trim={20px 32px 200px 120px}, clip, width=0.2\linewidth]{style_transfer_images/graph_images/A_400_0250.jpg}}
    
\end{tabular}

\vspace{-0.21cm}
\caption{\textbf{Tuning the objective at test-time:} First row shows the results of the fixed networks, each was trained separately for a different objective (corresponding to the red $\times$'s). Second row shows results for image interpolation between two fixed nets - $\lambda\!=\!2\cdot10^4$ and $\lambda\!=\!5\cdot10^5$ (corresponding to the green curve) as baseline. Third row shows results of Dynamic-Net with three tuning-blocks (main network was trained with $\lambda_0\!=\!2\cdot10^4$ and tuning-blocks with $\lambda_0\!=\!10^6$) where we increase $\alpha$ from 0 to 1 and $\alpha\!=\!\alpha_0\!=\!\alpha_1\!=\!\alpha_2$ (corresponding to the blue curve). The graph shows the ${\cal L}_{style}$ vs. ${\cal L}_{content}$ where the cyan dots represent a grid search of Dynamic-Net for 1000 possible values of $\alpha_0$, $\alpha_1$, $\alpha_2$. In the bottom right corner is a zoomed-in patch of our approach vs. image interpolation for $\alpha\!=\!0.25$, it can be observed that the baseline is dissolving one image upon the other reader than naturally increase the style as our approach and empirical evidence can be seen in the graph.}
\label{fig:style_vs_content}
\end{figure*}
%\input{figures/fig_lioness_style_transfer.tex}

%\paragraph{Fast Style Transfer}
\paragraph{Super-position of objectives:}
We begin with the common scenario where the objective-space is a super-position of two loss terms.
We followed the setup of \emph{fast style transfer}~\cite{johnson2016perceptual}, where the goal is to transfer the style of a specific style image to any input image. 
This is done by training a CNN to optimize the objective:
%\begin{equation}
${\cal O}\!= \!{\cal L}_{content}\!+\!\lambda {\cal L}_{style}$,
%\label{eq:style_obj}
%\end{equation}
where $ {\cal L}_{content} $ is the Perceptual loss~\cite{gatys2016image} between the output image and input image, and $ {\cal L}_{style} $ is the Gram loss~\cite{gatys2016image} between the output image and style image. The hyper-parameter $\lambda$ balances between preserving the content image and transferring the texture and appearance of the style image. 
Our goal here is to show that tuning $\alpha$ of the Dynamic-Net at test-time can replace tuning of $\lambda$ at training-time.

Following the training procedure suggested in Section~\ref{sec:proposed-approach} we first train the main network with objective ${\cal O}_0\!=\!{\cal L}_{content} + \lambda_0{\cal L}_{style}$, then freeze their weights and train the tuning-blocks with ${\cal O}_1\!=\!{\cal L}_{content} + \lambda_1 {\cal L}_{style}$.
%\footnote{$\lambda_0=2*10^4$, $\lambda_1=10^7$}. 
Similar to~\cite{johnson2016perceptual} we use the MS-COCO~\cite{lin2014microsoft} dataset for training. 
%\alon{In all the experiments in this section, If not stated otherwise, 3 tuning-blocks are used.} 

Figure~\ref{fig:style_vs_content} shows a few example results together with the corresponding working points in the objective-space, which trade-offs the content and style loss terms.
We successfully control the level of stylization, at test-time, by tuning $\alpha$.
An important result is that the working points emulated by the Dynamic-Net correspond to fixed networks trained for that specific working point (marked by $\times$) in terms of style loss and content loss.

The figure also compares to interpolation in image space, i.e., blending images produced by different fixed networks directly. For this baseline we use the following two networks; the main network of our Dynamic-Net and the closest fixed network (in terms of loss) to Dynamic-Net with $\alpha=1$. It can be seen that the results are inferior qualitatively and quantitatively, since the style loss does not change monotonically.

Our method also allow tuning each tuning-block individually as can be observed by the grid search in the graph. Each point of the grid search represent a result produced with a different value of $\alpha_0$, $\alpha_1$ and $\alpha_2$. This allows us to traverse the objective space in many interesting ways and even produce different images for the same working point.

%We further explore adding two sets of tuning blocks to a main network (optimized for $ \lambda $ that is between $ \lambda_0 $ and $ \lambda_1 $) and train every set to an opposite objective (first set optimized for $ \lambda_0 $ second set optimized for $ \lambda_1 $), thus allow interpolating between "closer" objective, this corresponds to the purple line in Figure~\ref{fig:lioness_style_transfer}. 

\begin{figure}[htbp!]
%\centering

\definecolor{lgreen}{RGB}{155,187,89}

\fboxsep=0mm%padding thickness
\fboxrule=1pt%border thickness

\setlength{\tabcolsep}{0pt}
\begin{tabular}{c c c c c}
    $\alpha\!=\!0$ & 
    $\alpha\!=\!0.25$ &
    $\alpha\!=\!0.5$ &
    $\alpha\!=\!0.75$ & 
    $\alpha\!=\!1$
    \\
    \rotatebox{90}{\hspace{6.3mm}\tiny{AdaIn \textcolor{white}{p}}}
    \begin{overpic}[trim={0 10px 0 20px}, clip, width=0.194\linewidth]{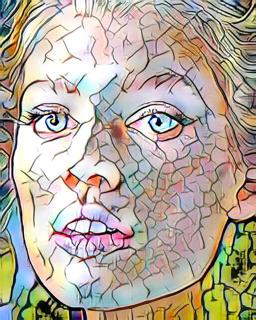}
        \put(3,3){\fcolorbox{lgreen}{lgreen}{\includegraphics[width=0.1\linewidth]{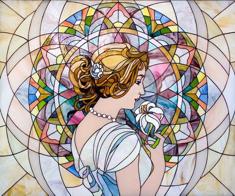}}}
    \end{overpic} &
    \includegraphics[trim={0 10px 0 20px}, clip,width=0.194\linewidth]{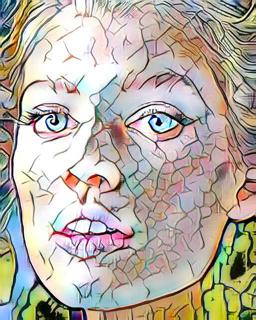} &
    \includegraphics[trim={0 10px 0 20px}, clip,width=0.194\linewidth]{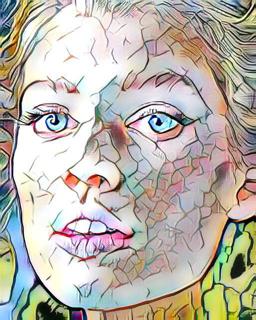} &
    \includegraphics[trim={0 10px 0 20px}, clip,width=0.194\linewidth]{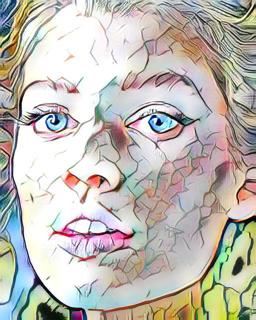} &
    \begin{overpic}[trim={0 10px 0 20px}, clip, width=0.194\linewidth]{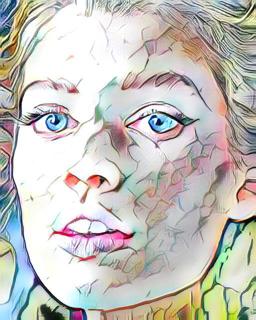}
        \put(34,3){\fcolorbox{lgreen}{lgreen}{\includegraphics[width=0.1\linewidth]{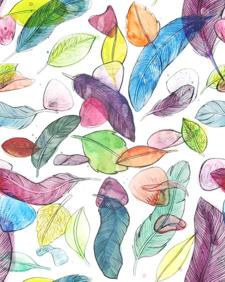}}}
    \end{overpic}
    \\
     \rotatebox{90}{\hspace{5.4mm}\tiny{Conditional IN \textcolor{white}{p}}}
    \includegraphics[trim={0 10px 0 20px}, clip,width=0.194\linewidth]{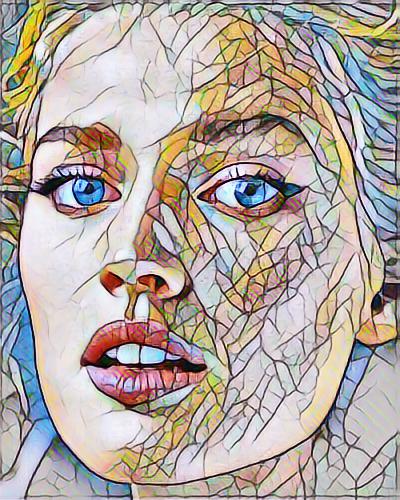} &
    \includegraphics[trim={0 10px 0 20px}, clip,width=0.194\linewidth]{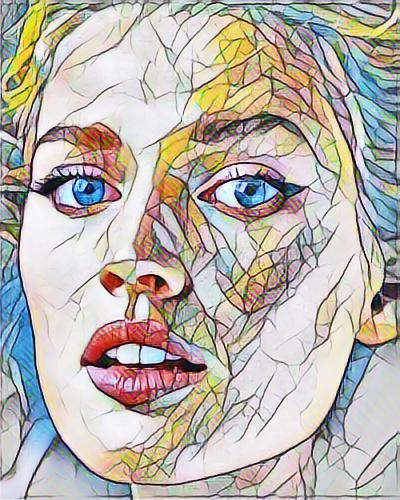} &
    \includegraphics[trim={0 10px 0 20px}, clip,width=0.194\linewidth]{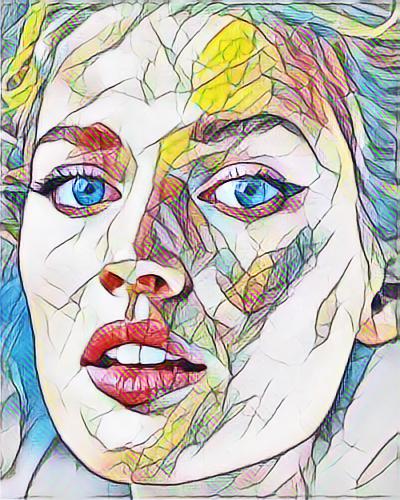} &
    \includegraphics[trim={0 10px 0 20px}, clip,width=0.194\linewidth]{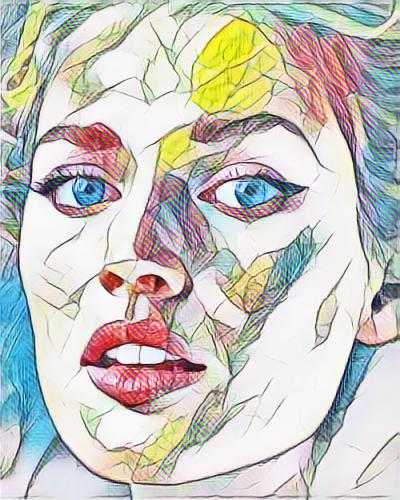} &
    \includegraphics[trim={0 10px 0 20px}, clip,width=0.194\linewidth]{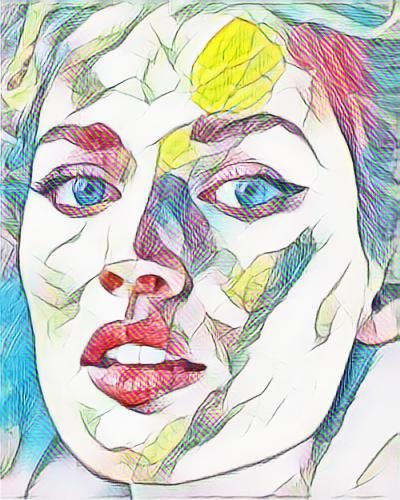}
    \\
     \rotatebox{90}{\hspace{5.3mm}\tiny{Image interp}}
    \includegraphics[trim={0 10px 0 20px}, clip,width=0.194\linewidth]{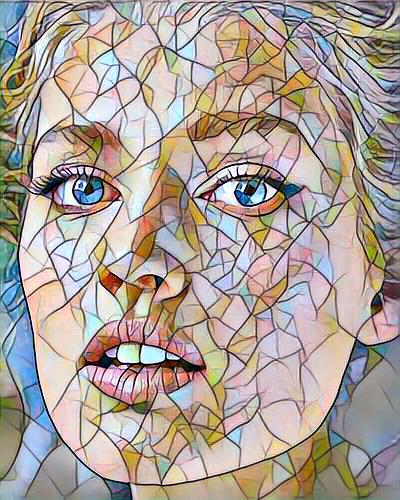} &
    \includegraphics[trim={0 10px 0 20px}, clip,width=0.194\linewidth]{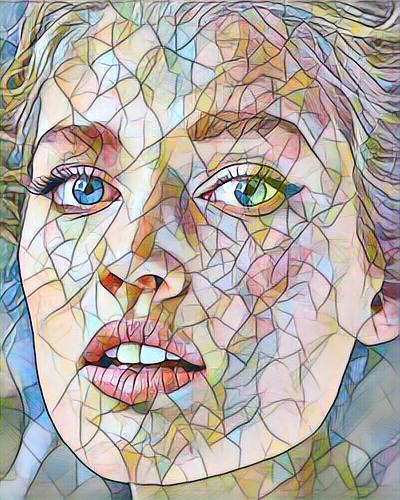} &
    \includegraphics[trim={0 10px 0 20px}, clip,width=0.194\linewidth]{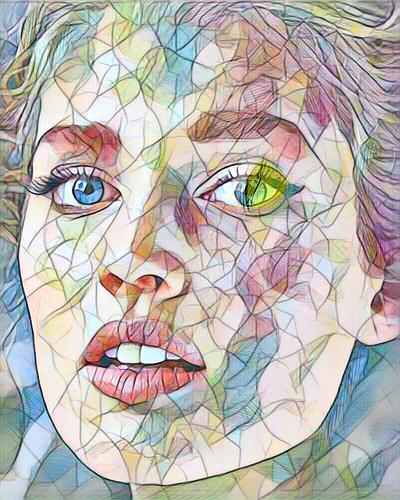} &
    \includegraphics[trim={0 10px 0 20px}, clip,width=0.194\linewidth]{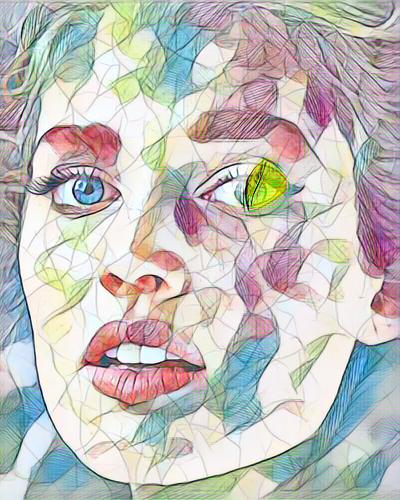} &
    \includegraphics[trim={0 10px 0 20px}, clip,width=0.194\linewidth]{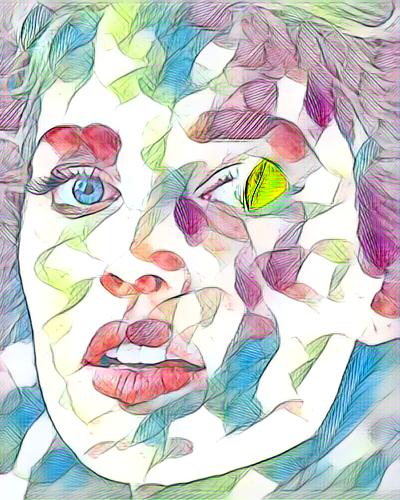}
    \\
    \rotatebox{90}{\hspace{6.3mm}\tiny{Ours \textcolor{white}{p}}}
    \includegraphics[trim={0 10px 0 20px}, clip,width=0.194\linewidth]{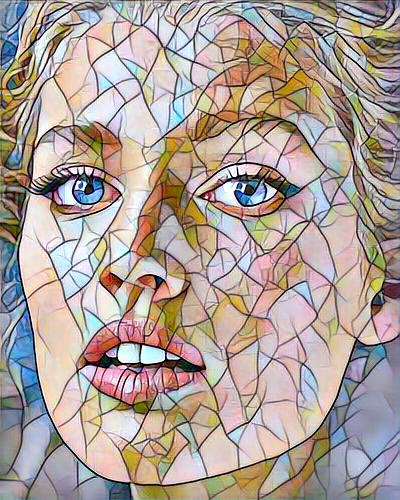} &
    \includegraphics[trim={0 10px 0 20px}, clip,width=0.194\linewidth]{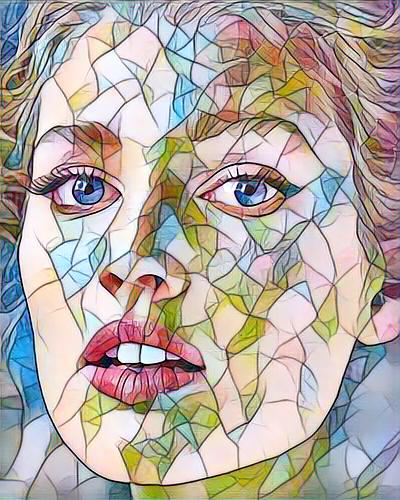} &
    \includegraphics[trim={0 10px 0 20px}, clip,width=0.194\linewidth]{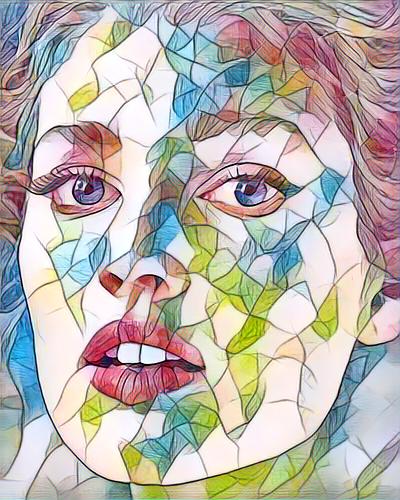} &
    \includegraphics[trim={0 10px 0 20px}, clip,width=0.194\linewidth]{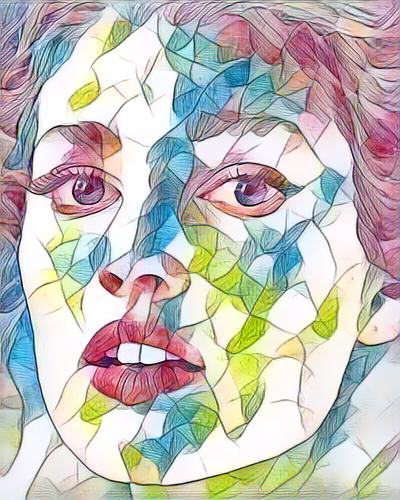} &
    \includegraphics[trim={0 10px 0 20px}, clip,width=0.194\linewidth]{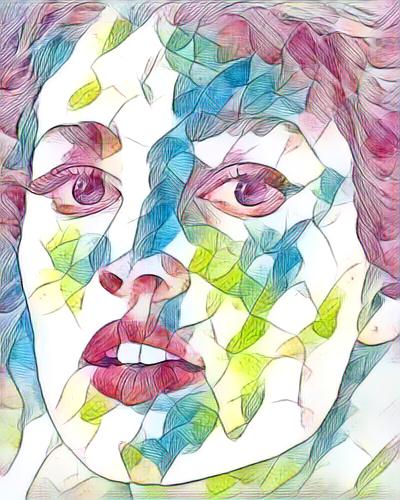}
\end{tabular}

\begin{tabular}{c c c}
    $\alpha\!=\!0$ & 
    %$\alpha\!=\!0.25$ &
    $\alpha\!=\!0.5$ &
    %$\alpha\!=\!0.75$ & 
    $\alpha\!=\!1$
    \\
    \rotatebox{90}{\hspace{7mm}\tiny{AdaIn \textcolor{white}{p}}}
    \begin{overpic}[trim={0 20px 0 20px}, clip, width=0.32\linewidth]{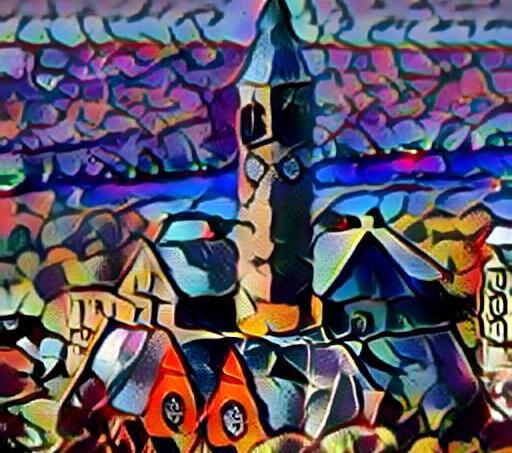}
        \put(3,3){\fcolorbox{lgreen}{lgreen}{\includegraphics[width=0.1\linewidth]{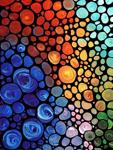}}}
    \end{overpic} &
    \includegraphics[trim={0 20px 0 20px}, clip, width=0.32\linewidth]{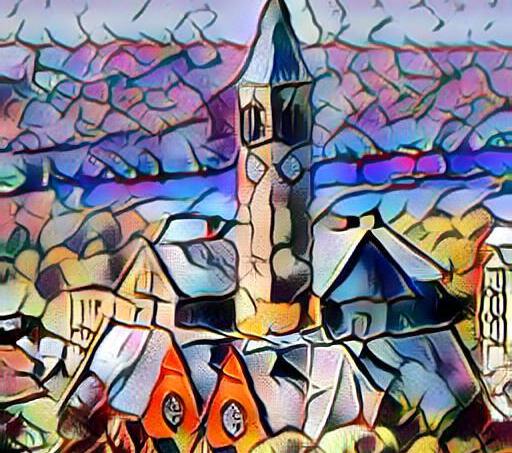} &
    \begin{overpic}[trim={0 20px 0 20px}, clip, width=0.32\linewidth]{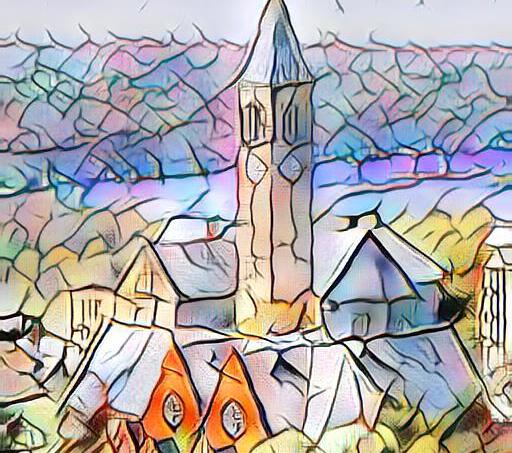}
        \put(65,3){\fcolorbox{lgreen}{lgreen}{\includegraphics[width=0.1\linewidth]{comp_images/mosaic.jpg}}}
    \end{overpic}
    \\
    \rotatebox{90}{\hspace{5.6mm}\tiny{Conditional IN \textcolor{white}{p}}}
    \includegraphics[trim={0 20px 0 20px}, clip, width=0.32\linewidth]{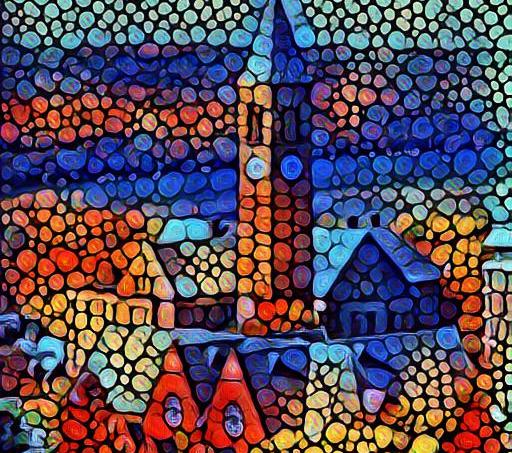} &
    \includegraphics[trim={0 20px 0 20px}, clip, width=0.32\linewidth]{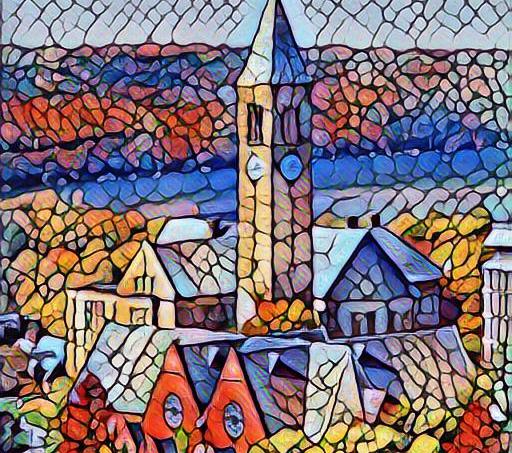} &
    \includegraphics[trim={0 20px 0 20px}, clip, width=0.32\linewidth]{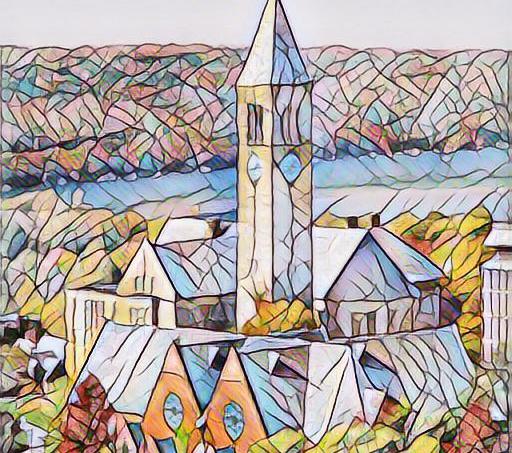}
    \\
    \rotatebox{90}{\hspace{5.4mm}\tiny{Image interp}}
    \includegraphics[trim={0 20px 0 20px}, clip, width=0.32\linewidth]{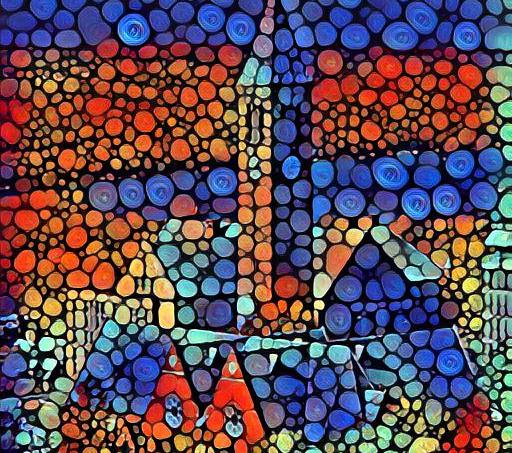} &
    \includegraphics[trim={0 20px 0 20px}, clip, width=0.32\linewidth]{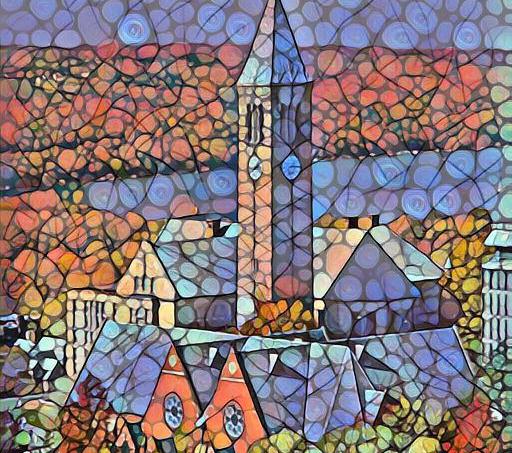} &
    \includegraphics[trim={0 20px 0 20px}, clip, width=0.32\linewidth]{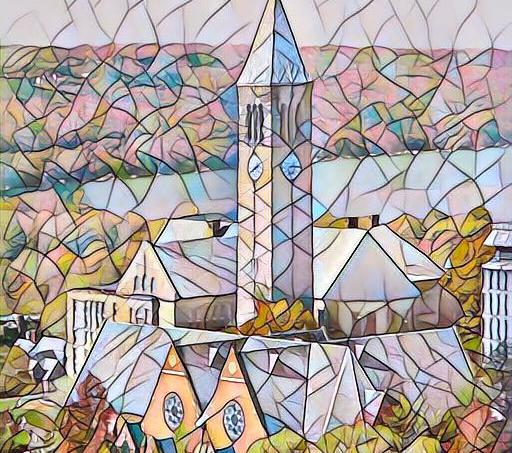}
    \\
    \rotatebox{90}{\hspace{7mm}\tiny{Ours \textcolor{white}{p}}}
    \includegraphics[trim={0 20px 0 20px}, clip, width=0.32\linewidth]{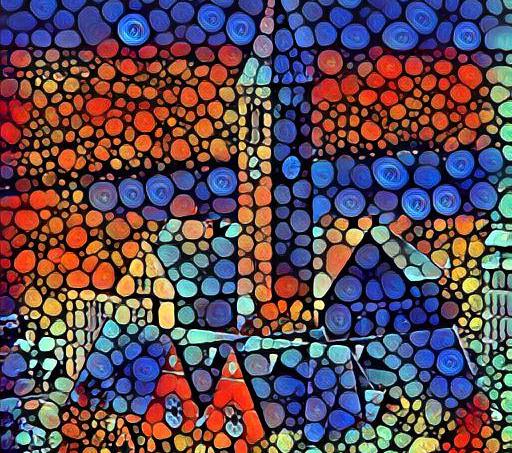} &
    \includegraphics[trim={0 20px 0 20px}, clip, width=0.32\linewidth]{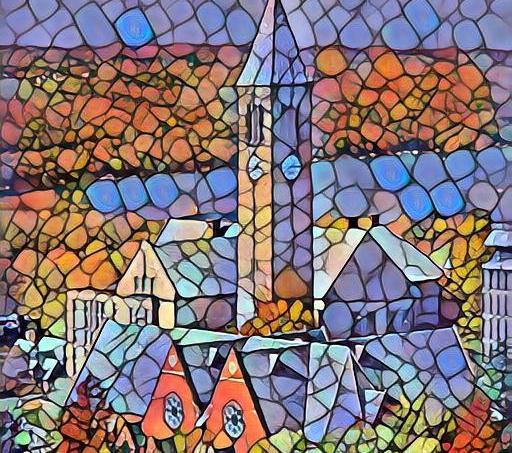} &
    \includegraphics[trim={0 20px 0 20px}, clip, width=0.32\linewidth]{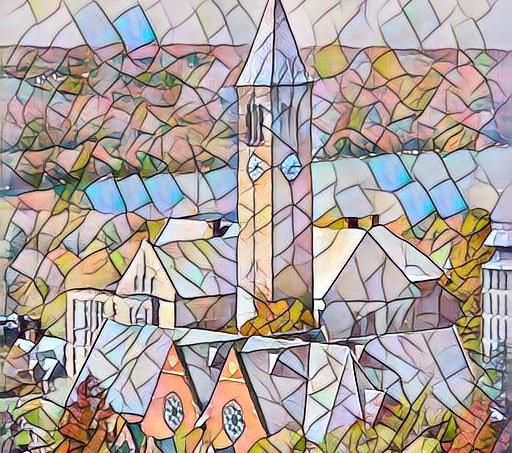}
    \\
\end{tabular}

\setlength{\tabcolsep}{1pt}
\begin{tabular}{c c c c}
    Ours & 
    Image interp &
    Ours & 
    Image interp 
    \\
    \includegraphics[width=0.24\linewidth]{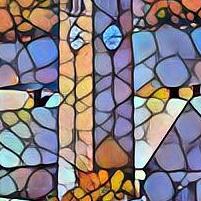} &
    \includegraphics[width=0.24\linewidth]{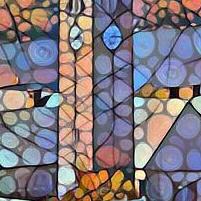} &
    \includegraphics[width=0.24\linewidth]{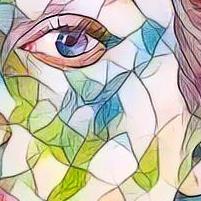} &
    \includegraphics[width=0.24\linewidth]{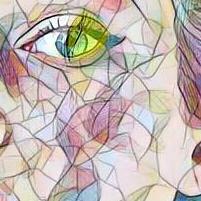}
\end{tabular}

\vspace{-0.2cm}
\caption{\textbf{Traversing between styles:} results of four different methods. Last row shows a zoomed-in patch of our method vs. image interpolation for $\alpha\!=\!0.5$. Best viewed zoomed-in.}
\label{fig:style_transfer_comparison}
\end{figure}
\begin{figure}
\footnotesize{
\begin{tabular}{p{0.05cm} p{1cm} p{1.1cm} p{0.9cm} p{1.3cm} p{1cm} p{1cm} p{0.8cm}}
    & $ \alpha\!=\!-1 $ & $ \alpha\!=\!-\frac{1}{2} $ & $ \alpha\!=\!0 $ & $ \alpha\!=\!0.5 $ & $ \alpha\!=\!1 $ & $ \alpha\!=\!\frac{3}{2} $ & $ \alpha\!=\!2 $
\end{tabular}
}
\includegraphics[width=0.99\linewidth]{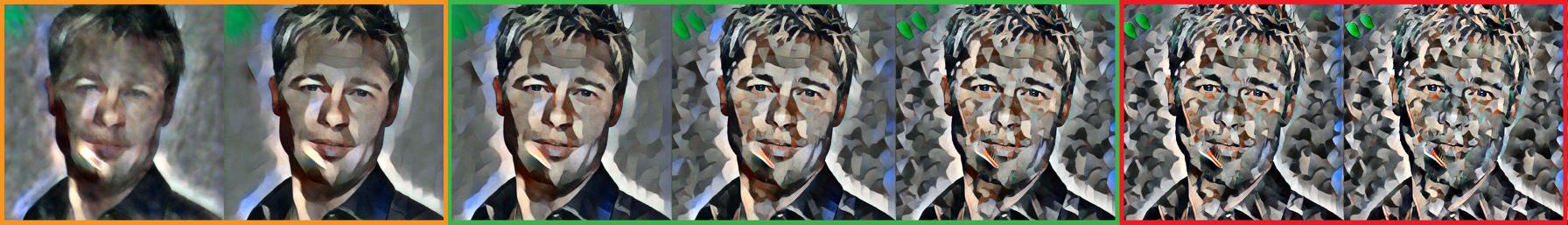}
\includegraphics[width=0.99\linewidth]{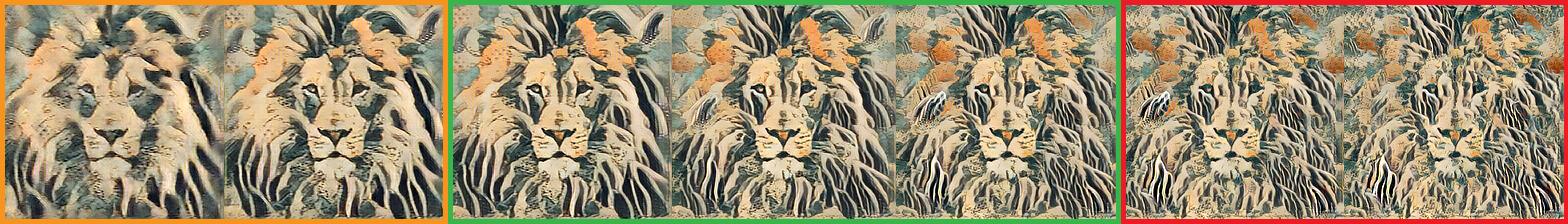}
\caption{\textbf{Objective extrapolation:} Style transfer results where the main-blocks are trained with a \textit{{high} resolution} style image, while the tuning-blocks are trained with a \textit{{low} resolution} style image. Since the tuning-blocks capture the trend between the two style images the Dynamic-Net can generate different scales of texture: (green-box) interpolation along the style scale (red-box): extrapolation to the low-resolution side, and (yellow): extrapolation to the high-resolution side (style images in the supplementary).}
\label{fig:style_transfer_extrapolation}
\end{figure}

\paragraph{Disjoint objectives:}
To further explore the generality of our approach we next experiment with disjoint objectives.
As a case study we chose to traverse between stylization with two different style images.
That is, ${\cal O}_0$ was trained with one style image, while ${\cal O}_1$ was trained with a different style image.
At test time we tune $\alpha$ to traverse between the two objectives.
Figure~\ref{fig:style_transfer_comparison} presents results when the style images are completely different. We compare our result to two algorithms, Arbitrary Style Transfer using AdaIN~\cite{Huang2017ArbitraryST} and Conditional-IN~\cite{dumoulin2017learned}. A third baseline is a simple interpolation in image space between results of two fixed networks. Conditional-IN reduces each style image into a point in an embedding space, each style shears the same convolutional weights of the network but it has its own normalization parameters. A blending effect is achieved by interpolating the normalization parameters of two styles.
Arbitrary Style Transfer network consists of a fixed-encoder, AdaIN layer and a decoder. AdaIN is used to adjust the mean and variance of the content input image to match those of an arbitrary style image. Interpolating between AdaIN parameters of two styles produces a blending effect. 
For AdaIN we use the official implementation and pre-trained network, while for Conditional-IN we use the official implementation but trained the network for 10 different styles used in our paper. For the image interpolation baseline we use the following two networks; the main network of our Dynamic-Net and a fixed network trained for the second style image.
Note that both AdaIN and Conditional-IN require specific constraints on the architecture, while our method is general and can extend existing high-quality pre-trained networks (as main networks). This can explain why our results are more faithful to the given style images. %\alon{is this last sentence ok?}
%Our results, that are more faithful to the style images, compared to AdaIN and Conditional-IN can be explained by the use of constraints on the architecture and training procedure that both method use in order to be implemented while our method is general and extends an existing successfully trained network (the main network).
In addition, as we can observe from the zoomed-in images, our method achieves a natural blending of the two styles as opposed to a "fade away" effect of one image on top of another, formed by the baseline.

In Figure~\ref{fig:style_transfer_extrapolation} the style images are two versions of the same style image, albeit at different resolutions.
It can be seen that Dynamic-Net provides a smooth transition between the objectives. 
We also examine the ability to \emph{extrapolate} in objective-space as shown in Figure~\ref{fig:style_transfer_extrapolation}.
Specifically, we wanted to see if we can emulate working points that are not intermediate to those used during training.
Interestingly, setting $\alpha<0$ or $\alpha>1$ also leads to meaningful results, corresponding to extrapolation in scale space of the style.

\begin{figure}
\centering
  \setlength{\tabcolsep}{0pt}
\begin{tabular}{c}
\textbf{Male  $\xrightarrow[]{}$ Female}\\
\includegraphics[width=\linewidth]{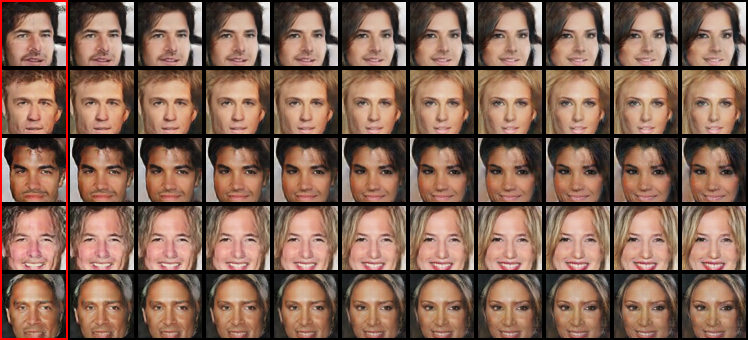}\\
\textbf{Dark Hair $\xrightarrow[]{}$ Blond Hair}\\
\includegraphics[width=\linewidth]{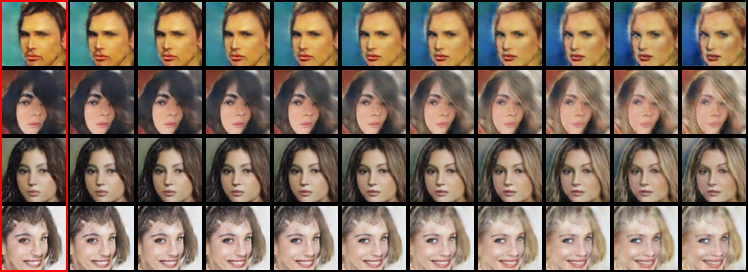}\\
$\alpha\!=\!0\xRightarrow{\hspace*{6.5cm}} \alpha\!=\!1$
\end{tabular}
%\vspace{-0.2cm}
\caption{\textbf{Controlled generation results:} the proposed method allow us to interpolate between different facial attributes. The values of $\alpha$ are gradually increasing from left to right, results in a monotonic change of the specific attribute. Most left: $\alpha\!=\!0$ correspond to the baseline result of DCGAN~\cite{radford2015unsupervised}.}
\label{fig:dcgan_results}
\end{figure}

\subsection{The objective is user specific\\:: Face Generation}
\vspace{-0.1cm}
In some applications the desired output is not only image dependent but further depends on the user's preference. 
This could be observed previously in the style transfer experiments, were every user could prefer different stylization options.
As another example for such a case we chose the task of face generation, where our approach endows the user with fine control over certain facial attributes, such as hair color or gender.

%\paragraph{:: Face Generation}
%\label{sec:generation}
We adopted the architecture of DCGAN~\cite{radford2015unsupervised} that is trained with a single adversarial loss ${\cal O}={\cal L}_{adv}$ over the CelebA~\cite{liu2015faceattributes} dataset.
To provide control over an attribute, such as hair color, we split the dataset into two sub-sets, e.g., dark hair vs. blond.
Both, the main network and the tuning-blocks were trained with an adversarial loss, but with different data sub-set.
% The main network was trained with an adversarial loss over one sub-set, while a single tuning-block was trained with an adversarial loss over the other sub-set.
The two objectives are thus disjoint in this case.

At test-time, the user can tune $\alpha$ to generate a face with desired properties. For example, the user can tune the hair color or the masculinity of the generated face.
Qualitative results are presented in Figure~\ref{fig:dcgan_results} for two attributes: male-to-female and dark hair-to-blond hair.
It can be seen that our Dynamic-Net smoothly traverses between the two objectives, generating plausible images, with a smooth attribute control.

%%%%%%%%%%%%%%%%%%%%%%%%%%%%%%%%%%%%%%%%%%%%%%%%%%%%%%%%%%%%
\begin{figure}[t]
\centering
\centering
\setlength{\tabcolsep}{1pt}
\begin{tabular}{c | c}
    Input & Network results \\
    \hline
    % image 510
    \begin{tabular}{l}
        \\
        \includegraphics[width=0.23\linewidth]{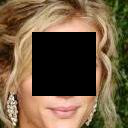}
    \end{tabular}
    &
    \setlength{\tabcolsep}{0pt}
    \begin{tabular}{c c c}
        $\alpha\!=\!0$ & $\alpha\!=\!0.4$ & $\alpha\!=\!1$ \\
        \includegraphics[width=0.23\linewidth]{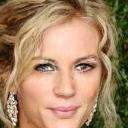} &
        \includegraphics[width=0.23\linewidth]{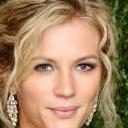} &
        \includegraphics[width=0.23\linewidth]{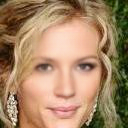} \\
    \end{tabular}
    \\
    % image 662
    \begin{tabular}{l}
        \\
        \includegraphics[width=0.23\linewidth]{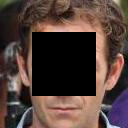}
    \end{tabular}
    &
    \setlength{\tabcolsep}{0pt}
    \begin{tabular}{c c c}
        $\alpha\!=\!0$ & $\alpha\!=\!0.4$ & $\alpha\!=\!1$ \\
        \includegraphics[width=0.23\linewidth]{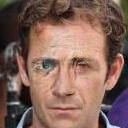} &
        \includegraphics[width=0.23\linewidth]{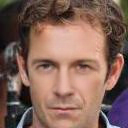} &
        \includegraphics[width=0.23\linewidth]{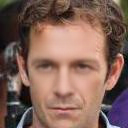} \\
    \end{tabular}
    \\
    % image 107
    \begin{tabular}{l}
        \\
        \includegraphics[width=0.23\linewidth]{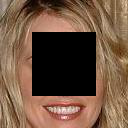}
    \end{tabular}
    &
    \setlength{\tabcolsep}{0pt}
    \begin{tabular}{c c c}
        $\alpha\!=\!0$ & $\alpha\!=\!0.4$ & $\alpha\!=\!1$ \\
        \includegraphics[width=0.23\linewidth]{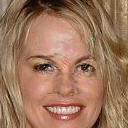} &
        \includegraphics[width=0.23\linewidth]{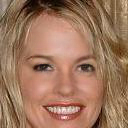} &
        \includegraphics[width=0.23\linewidth]{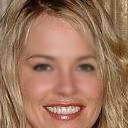} \\
    \end{tabular}
    \\
    % image 88
    % \begin{tabular}{l}
    %     \\
    %     \includegraphics[width=0.23\linewidth]{inpainting_images2/input_000087.png}
    % \end{tabular}
    % &
    % \setlength{\tabcolsep}{0pt}
    % \begin{tabular}{c c c}
    %     $\alpha\!=\!0$ & $\alpha\!=\!0.5$ & $\alpha\!=\!1$ \\
    %     \includegraphics[width=0.23\linewidth]{inpainting_images2/05_0088_0_0.png} &
    %     \includegraphics[width=0.23\linewidth]{inpainting_images2/05_0088_0_10.png} &
    %     \includegraphics[width=0.23\linewidth]{inpainting_images2/05_0088_1_20.png} \\
    % \end{tabular}
    % \\
    % image 215
    \begin{tabular}{l}
        \\
        \includegraphics[width=0.23\linewidth]{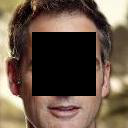}
    \end{tabular}
    &
    \setlength{\tabcolsep}{0pt}
    \begin{tabular}{c c c}
        $\alpha\!=\!0$ & $\alpha\!=\!0.5$ & $\alpha\!=\!1$ \\
        \includegraphics[width=0.23\linewidth]{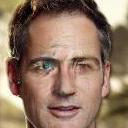} &
        \includegraphics[width=0.23\linewidth]{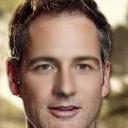} &
        \includegraphics[width=0.23\linewidth]{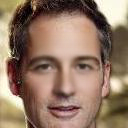} \\
    \end{tabular}
    \\
    % image 632
    \begin{tabular}{l}
        \\
        \includegraphics[width=0.23\linewidth]{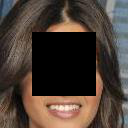}
    \end{tabular}
    &
    \setlength{\tabcolsep}{0pt}
    \begin{tabular}{c c c}
        $\alpha\!=\!0$ & $\alpha\!=\!0.4$ & $\alpha\!=\!1$ \\
        \includegraphics[width=0.23\linewidth]{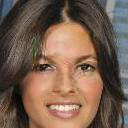} &
        \includegraphics[width=0.23\linewidth]{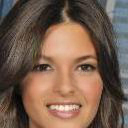} &
        \includegraphics[width=0.23\linewidth]{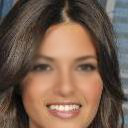} \\
    \end{tabular}
    \\

\end{tabular}
%\vspace{-0.21cm}
\caption{\textbf{Robustness to hyper-parameter:} The main network ($\alpha\!=\!0$) produces artifacts common to adversarial training while $\alpha\!=\!1$ produces blurry images common to L1 loss. Using $0\!<\!\alpha\!<\!1$ results in high quality images preventing the need to retrain the main network numerous times with different objectives to achieve high quality results.}
\label{fig:inpainting_only_net2}
\end{figure}
\subsection{Robustness to hyper-parameter\\:: Image Completion}
\vspace{-0.1cm}
%\todo{\textbf{Robustness:} all applications, image completion - robustness to objective.}

Our last goal is to show that our approach shrinks the required search space over the objective at training time.
In this experiment we use the task of image completion to show that by using tuning-blocks we can effectively prevent exhaustive re-training for tuning the objective. We intentionally train the main network for a sub-optimal objective that leads to poor quality completion and artifacts and use the tuning-blocks to adjust the objective post training to achieve high quality results.
This was in order to show that even when the main network is of poor quality, adding the tuning-blocks with an appropriate $\alpha$ could result in a better overall network, getting rid of the artifacts.
This is possible because we can traverse the objective space and thus identify good working points, even when those used for training were sub-optimal.

In our experimental setup the input is a face image with a large hole at the center, and the goal is to complete the missing details in a faithful and realistic manner.
As architecture we adopted a version of \textit{pix2pix}~\cite{isola2017image} (see supplementary for details).
The objective for training the main network was ${\cal O}_0 = {\cal L}_{L1} + \lambda {\cal L}_{adv}$ ($\lambda = 0.005$) and three tuning-blocks were trained with ${\cal O}_1 = {\cal L}_{L1}$.

Figure~\ref{fig:inpainting_only_net2} shows some of our results. The main network ($\alpha=0$) produces artifacts common for adversarial losses while on the other hand using the whole wight of the tuning-blocks ($\alpha=1$) results in blurry images common to L1 losses. Using $0<\alpha<1$ We traverse the objective space and produce high quality images.
This suggests that during training rather then trying multiple values for $\lambda$ one can just select a single value, and then at test-time adapt $\alpha$.
The training of the tuning-blocks demonstrate robustness and implies that our Dynamic-Net forms a good alternative to the traditional greedy search. Choosing $\alpha$ can be done interactively in real-time, to tailor the network for a specific image. Since for different images there can be found a better objective that suit them specifically, interactively editing the results per image can be significant and difficult to achieve using fixed networks.   
Setting $\alpha$ is fast and provides an interesting alternative to hyper-parameter search at training time, both in terms of computing efficiency and as it enables image and user specific tuning.

\begin{figure}
\centering
%\includegraphics[width=0.7\linewidth]{style_transfer_images/fail_case/fail_case_v2.png}
%\includegraphics[width=0.7\linewidth]{style_transfer_images/fail_case/fail_case_v3.png}
%\bgroup
%\def\arraystretch{0}
\fboxsep=0mm%padding thickness
\fboxrule=1pt%border thickness

%\definecolor{my_blue}{RGB}{91,155,213}
%\definecolor{my_green}{RGB}{146,208,80}
\definecolor{fixed_nets}{RGB}{192,80,77}
\definecolor{ours}{RGB}{79,129,189}
\definecolor{base_line}{RGB}{155,187,89}

\includegraphics[width=1\linewidth]{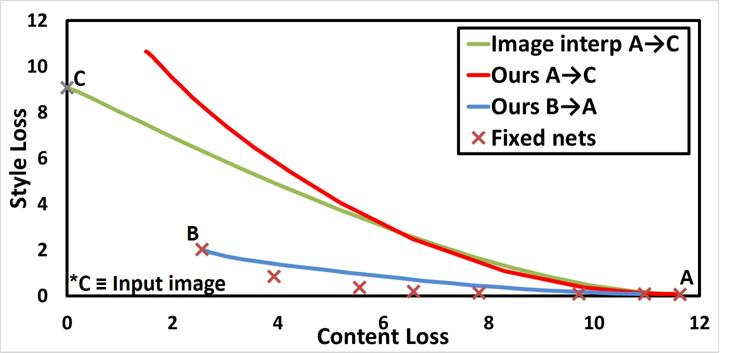}
\vspace{-0.15cm}

\setlength{\tabcolsep}{0pt}
\begin{tabular}{c c c c}
     \fcolorbox{base_line}{base_line}{\begin{overpic}[width=0.24\linewidth]{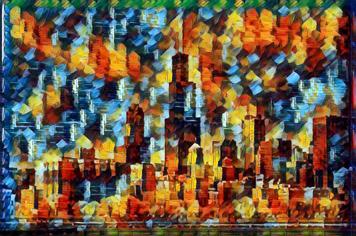}
        \put(3,3){\fcolorbox{base_line}{base_line}{\includegraphics[width=0.1\linewidth]{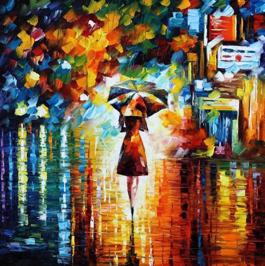}}}
     \end{overpic}} &
     %\fcolorbox{my_blue}{my_blue}{\includegraphics[width=0.24\linewidth]{style_transfer_images/fail_case/input_0000.jpg}} & 
     \fcolorbox{base_line}{base_line}{\includegraphics[width=0.24\linewidth]{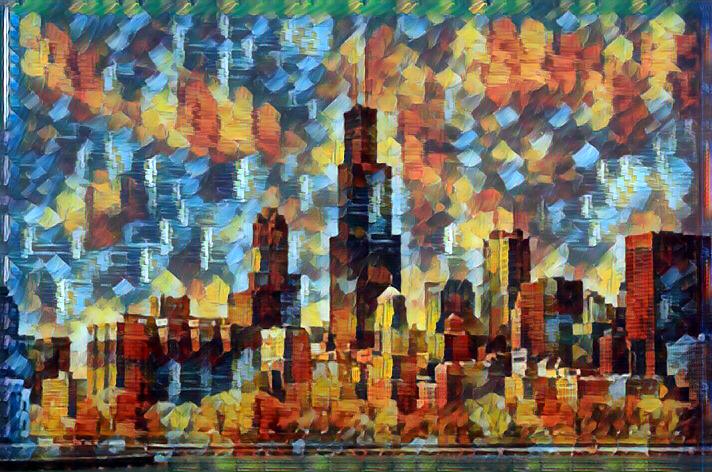}} & 
     \fcolorbox{base_line}{base_line}{\includegraphics[width=0.24\linewidth]{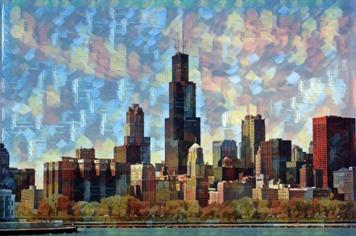}} & 
     \fcolorbox{base_line}{base_line}{\includegraphics[width=0.24\linewidth]{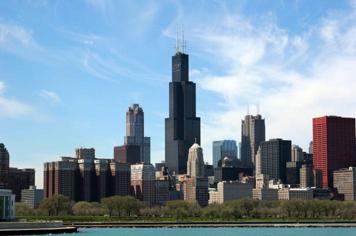}}
     \\
%\end{tabular}
%\setlength{\tabcolsep}{0pt}
%\begin{tabular}{c c c c}
     \fcolorbox{red}{red}{\includegraphics[width=0.24\linewidth]{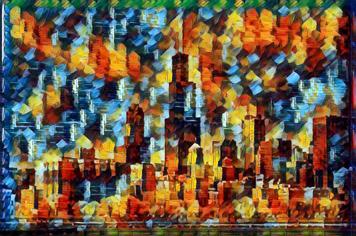}} & 
     \fcolorbox{red}{red}{\includegraphics[width=0.24\linewidth]{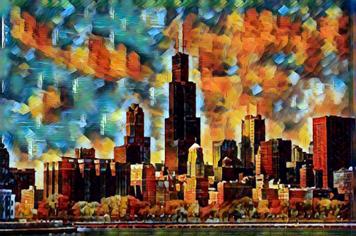}} & 
     \fcolorbox{red}{red}{\includegraphics[width=0.24\linewidth]{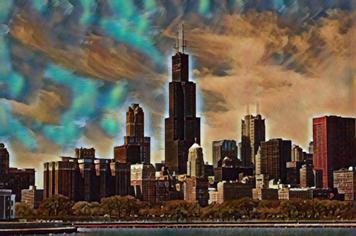}} & 
     \fcolorbox{red}{red}{\includegraphics[width=0.24\linewidth]{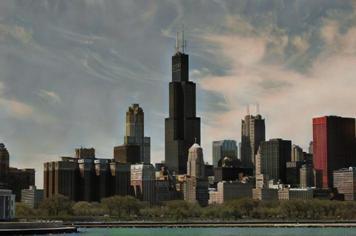}}
     \\
%\end{tabular}
%\setlength{\tabcolsep}{0pt}
%\begin{tabular}{c c c c}
     \fcolorbox{ours}{ours}{\includegraphics[width=0.24\linewidth]{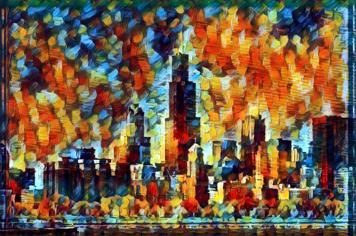}} & 
     \fcolorbox{ours}{ours}{\includegraphics[width=0.24\linewidth]{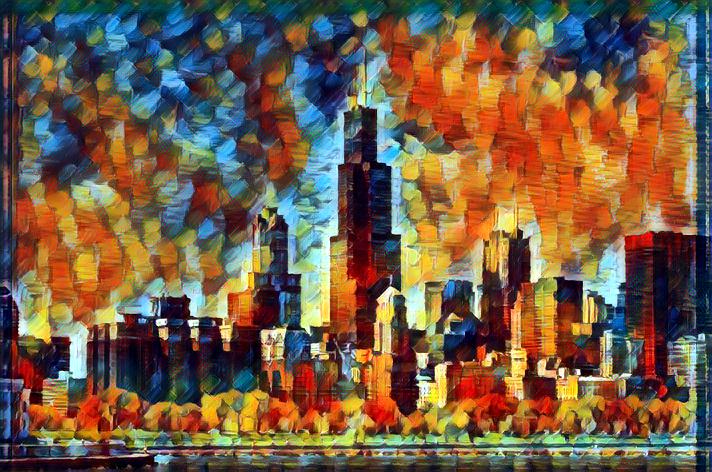}} & 
     \fcolorbox{ours}{ours}{\includegraphics[width=0.24\linewidth]{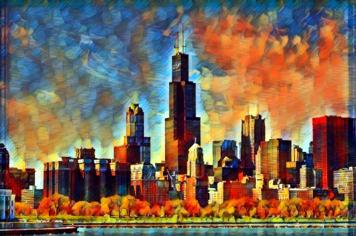}} & 
     \fcolorbox{ours}{ours}{\includegraphics[width=0.24\linewidth]{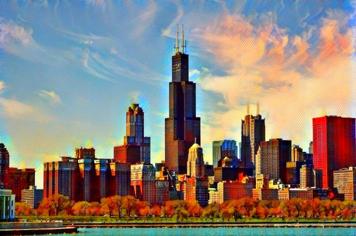}}
\end{tabular}

\setlength{\tabcolsep}{2pt}
\begin{tabular}{c c}
    %Image interp ($\alpha\!=!\0.5$) & Ours ($\alpha\!=!\0.5$)
    Image interp (2nd column) & Ours (2nd column)
    \\
    \fcolorbox{base_line}{base_line}{\includegraphics[trim={30px 100px 200px 50px}, clip, width=0.48\linewidth]{style_transfer_images/fail_case/input_0250.jpg}} &
    \fcolorbox{ours}{ours}{\includegraphics[trim={30px 100px 200px 50px}, clip, width=0.48\linewidth]{style_transfer_images/fail_case/01_0750.jpg}}
\end{tabular}
%\egroup
% \includegraphics[width=0.9\linewidth]{style_transfer_images/fail_case/chicago_blue_fail_case.png}
% \includegraphics[width=0.9\linewidth]{style_transfer_images/fail_case/chicago_red_fail_case.png}
% \includegraphics[width=0.9\linewidth]{style_transfer_images/fail_case/chicago_green_fail_case.png}
%\vspace{-0.3cm}
\caption{\textbf{Failure Case:} \textbf{Top}: Each curve corresponds to a different setting: 
(green) image-space interpolation between fixed net (A) and the input image.
(red) Dynamic-Net results with ${\cal O}_0\!=\!{\cal O}_A$ ($\lambda\!=\!10^6$) and ${\cal O}_1\!=\!{\cal O}_C$.
(blue) Dynamic-Net results with ${\cal O}_0\!=\!{\cal O}_B$ ($\lambda\!=\!10^4$) and ${\cal O}_1\!=\!{\cal O}_A$. 
\textbf{Bottom}: Example images, the box color correspond to the curve color.
Training with medium objective range ($B\!\to\!A$) achieved great results, however, increasing the range too much, i.e. $A\!\to\!C$, weaken the results quality.
}
\label{fig:style_transfer_fail_case}
\end{figure}
\section{Method Analysis}
\vspace{-0.1cm}
%We next present two additional experiments on the task of style transfer, the first compare the two suggested frameworks and the second discuss the method limitations.
\label{sec:analysis}
%\input{figures/fig_single_vs_multi_block.tex}
%\input{figures/fig_multi_net.tex}

% \textbf{Single-block vs. Multi-block \ \ }
% \label{sec:multi_single}
% Figure~\ref{fig:single_vs_multi_block} compares between our two frameworks. As the plot shows, the single-block framework (blue and purple curves) produce a curve in the objective space, while the multi-block framework produce an entire surface over the objective space (green dots).   
% Furthermore, the single-block framework emulates the fixed-nets slightly better, however, the multi-block framework better minimize objective ${\cal O}_1$ (point B).

\textbf{Limitations \  \ }
\label{sec:limitations}
Figure~\ref{fig:style_transfer_fail_case} present the limitation of the proposed method, when using extreme objectives. Specifically, we trained the tuning-blocks \emph{without} a style loss term, i.e. ${\cal O}_1\!=\!{\cal L}_{content}$. 
We observe that, the simple image interpolation (green curve) achieves better results than our method (red curve) when approaching near point C, that is, near the objective ${\cal O}_1$.
The main reason for that, is that the main network was trained for style transfer, and the ability of the tuning blocks to ``Turn the table upside down'' and produce image with very little style, is limited.
Last, we show that using Dynamic-Net with a smaller range between the objectives, $B\!\to\!C$, (blue curve) outperform both methods and approximate the fixed nets accurately.

\section{Conclusions }
\label{sec:conclusions}
\vspace{-0.1cm}
We propose Dynamic-Net a novel two phase training framework that allow traversing the objective space at inference time without re-training the model. We have shown its broad applicability on variety vision tasks: style transfer, face generation and image completion. In all application we showed that our method allow easy and intuitive control of the objective trade-off. This work is a first step in providing a model that is not limited to a specific static working point -- a dynamic model. Future work include bringing the dynamic concept to other application and expend it to other objective spaces.

In the supplementary we present additional results and provide implementation details.

\section*{Acknowledgements }
This research was supported by the Israel Science Foundation under Grant 1089/16 and by the Ollendorf foundation.

\clearpage
{\small
\bibliographystyle{ieee_fullname}
\bibliography{mybib}
}

\end{document}